\documentclass[conference]{IEEEtran}
\IEEEoverridecommandlockouts
\usepackage[american]{babel}
\usepackage{cite}
\usepackage{amsmath,amssymb,amsfonts}
\usepackage{mathtools}
\usepackage{algorithmic}
\usepackage{graphicx}
\usepackage{textcomp}
\usepackage{xcolor}
\def\BibTeX{{\rm B\kern-.05em{\sc i\kern-.025em b}\kern-.08em
    T\kern-.1667em\lower.7ex\hbox{E}\kern-.125emX}}

\usepackage{url}
\usepackage{tikz}
\usepackage{microtype}

\usepackage[ansinew]{inputenc}
\usepackage{amstext}
\usepackage{amsthm}
\usepackage{cleveref}
\usepackage{csvsimple}

\usepackage{bbm}

\newcommand{\bigslant}[2]{{#1}/{#2}} 
\newtheorem{definition}{Definition}
\newtheorem{problem}{Problem}

\begin{document}

\title{{Distances for WiFi Based Topological\\ Indoor Mapping} 
\thanks{Funded by DFG SPP 1894:  Volunteered Geographic Information: Interpretation, Visualisierung und Social Computing.}
}


\author{\IEEEauthorblockN{1\textsuperscript{st} Bastian Schaefermeier}
\IEEEauthorblockA{\textit{L3S Research Center} \\
\textit{Leibniz University}\\
Hannover, Germany \\
schaefermeier@l3s.de}
\and
\IEEEauthorblockN{2\textsuperscript{nd} Tom Hanika}
\IEEEauthorblockA{\textit{Knowledge and Data Engineering Group} \\
\textit{University of Kassel}\\
Kassel, Germany \\
tom.hanika@cs.uni-kassel.de}
\and
\IEEEauthorblockN{3\textsuperscript{rd} Gerd Stumme}
\IEEEauthorblockA{\textit{Knowledge and Data Engineering Group} \\
\textit{University of Kassel}\\
Kassel, Germany \\
stumme@cs.uni-kassel.de}
}

%

\maketitle

\begin{abstract}
  For localization and mapping of indoor environments through WiFi
  signals, locations are often represented as likelihoods of the
  received signal strength indicator. In this work we compare various
  measures of distance between such likelihoods in combination with
  different methods for estimation and representation. In particular,
  we show that among the considered distance measures the Earth
  Mover's Distance seems the most beneficial for the localization
  task. Combined with kernel density estimation we were able to retain
  the topological structure of rooms in a real-world office scenario.
\end{abstract}


\section{Introduction}
Indoor localization and mapping through WiFi signals has received a
lot of interest in research through the recent years. This interest
has largely been accelerated through the prevalent use of smartphones,
as these allow for localization without the need of any specialized
additional hardware. Since nowadays WiFi access points (APs) are
present in most situations within buildings, no additional
infrastructure needs to be installed. One typical method for
localization and mapping in this realm is to compare the distributions
of received signal strength indicators (RSSIs). An advantage of this
method is that the APs do not need to be modified. Therefore
localization and mapping can be performed virtually everywhere.

A difficult problem that frequently arises in such scenarios is to
measure the (dis)similarity between two RSSI distributions. This
(dis)similarity then is interpreted as a (dis)similarity of locations
in the following way: similar measurements belong to nearby
locations. This notion of dissimilarity enables us to apply
unsupervised machine learning procedures, e.g., clustering, in order
to identify locations. Some techniques were already introduced to
measure (dis)similarities between RSSI distributions (e.g.,
\cite{ref_kl}). However, they were often used in supervised machine
learning scenarios. The focus there is not necessarily on quality of
the applied distance measure, but on learning through labeled
examples.
%

In this paper we investigate various distance measures between sets of
WiFi observations, where we also consider various representational
methods for the measurements. Since WiFi signals suffer from strong
variance even when the location is not changed, we especially consider
representations through discrete and continuous probability
distributions which allow for modeling the uncertainty in RSSI
observations.

In this work we focus on the following scenario. First, pedestrians move
indoors with a smartphone following their normal behaviour, e.g., at
the work place. Secondly, our aim is to infer sets of locations, i.e.,
we want to recognize from recorded WiFi signals whether some location
is revisited or whether a different location is visited. Thirdly, we
do not require that additional infrastructure or software is installed
at the
locations. 

Our contributions are as follows: (1) To the best of our knowledge,
this work is the first thorough study of various measures of
distance between RSSI distributions. (2) We describe a simple method
to group together RSSI measurements which are made at a single
location when a smartphones is not moved. (3) We substantiate our study
with a real world office scenario experiment over five days. Our
participants in this experiment simply follow their normal day
behaviour, while carrying a smartphone. This is in contrast to
research where participants are instructed to walk along predefined
paths or to hold their smartphone in a specific way.

%
\section{Problem Statement}
We will start by introducing the task and constraints in an informal way. After that we give a formal problem definition and finally, decompose the problem into feasible subproblems.
\subsection{Task and requirement definition}
Our ultimate aim is to compute topological maps from signal strength
measurements of WiFi access points (APs). The measurements are made
passively through people carrying smartphones while they follow their
normal behaviour. A topological map should reflect which physical
locations there are and how these locations are related to each
other. In this paper, we focus on the recognition and discrimination
of distinct physical locations through measured WiFi signal
strengths. Physical locations can, for example, be a kitchen where
people brew coffee or a canteen where people have lunch. In a more fine-grained scenario, physical locations could refer to the $\varepsilon$-neighborhood of points in a floor plan. In our experiments we will, however, only consider the case where locations are defined on the room level. 

We pose the following constraints on a solution for this task:
\begin{enumerate}
\item \emph{Independent}: Our method should work without any need of additional infrastructure at the physical locations to be mapped.
\item \emph{Automatic}: Our method should not require a specific user behaviour (as for example keeping the smartphone in your hand pointing into the walking direction).
\item \emph{Effortless}: No user interaction or manual place annotation should be needed.
\item \emph{Lightweight}: We aim at a lightweight approach, i.e., few data should be needed and, following the notion of Occam's razor, the model complexity should be low.
\end{enumerate}
In this work we consider a scenario where a set of distinct physical
locations should be recognized from a set of WiFi measurements. In
addition to that we want to assign these measurements to physical
locations in order to perform device localization. This should be
achieved without the help of any ground truth, i.e., through an
unsupervised
method. 

To discriminate sets of measurements, one needs, first, a
representation of the measurements (often called a \emph{fingerprint} of
a location), and secondly, a measure of dissimilarity or \emph{distance}
between the representations.
In supervised localization, RSSI measurements are often represented as
vectors where each component denotes the RSSI of a specific access
point measured at a point in time \cite{radar}. We, however, only
consider representations through probability distributions of RSSI
values. These distributions will be conditioned on
locations. Hence, we call those conditional distributions
\emph{RSSI likelihoods}. The reason for considering RSSI likelihoods
is that single RSSI measurements have strong random fluctuations, even
when the location is stable. A probability distribution can represent
these fluctuations. Additionally, probability distributions naturally
allow for further inferences, e.g., localization of previously unseen
measurements through maximum likelihood or determining a confidence of
being at some location (which is the reason for choosing the name RSSI
likelihoods).


We consider several methods to estimate and represent RSSI likelihoods
and various measures of \emph{distance} between them. While some authors
consider a distance measure as a synonym for a metric, we consider  in
this work a more general notion.
\begin{definition}[Distance measure on X]
  \label{def:distx}
A distance measure on a set $X$ is a function $d: X \times X
\rightarrow \mathbb{R}_{\ge 0}$ such that for all $a,b \in X$ we have $ a=b \implies d(a,b)=0$.
\end{definition}
Nonetheless, we will require that a distance measure gives an
interpretation of dissimilarity, where a higher distance is
interpretable as higher dissimilarity. 
%
Our aim  is to find the best combination of likelihood estimation
and distance measure, such that calculated distances between RSSI
likelihoods reflect real distances between the physical locations
appropriately. To evaluate particular combinations we consider
first,  properties of the distance measure, secondly, discriminative
strength, thirdly, correlations with $L_{2}$ distances in
a floor plan. The latter two will be evaluated through a real-world experiment. 

\subsection{Formal Problem Definition}
\label{sec:formalProb}

Our input data consists of timestamped WiFi observations made by
devices, e.g., smartphones, which we identify with a person
carrying it. A WiFi observation is the received signal strength
indicator (RSSI) measured by a device from an AP. We use RSSI
measurements, because of the relation between signal strengths
and physical distances to APs. Each AP is uniquely identified through
a basic service set identifier (BSSID).
\begin{definition}[WiFi data set]
We call the quaternary relation  \( W \subseteq \mathbb{N} \times D \times B
\times R \) \emph{WiFi data set}, where  $\mathbb{N}$ represents timestamps, $D$ is a set of devices, $B$ a set of BSSIDs and $R=[-100, -10]\cap \mathbb{Z}$ a range of RSSI values. A WiFi observation $o\coloneqq (t, d, b, r ) \in W$ contains the  RSSI value $r$ of WiFi access point $b$ sensed at timestamp $t$ by device $d$. For all $(t,d,b,s_1), (t,d,b,s_2) \in W$ it shall hold that $s_1 = s_2$.
\end{definition}


We require $\check{L}\subseteq W \times W$ to be an equivalence
relation (i.e., reflexive, symmetric and transitive relation) which
reflects the \textsl{true} association between observations and physical
locations. So, we suppose the existence of some ground
truth for distinct or equal physical locations in the WiFi
observations. Hence, this requires that for all
$(t,d,b_1, r_1) \in W \text{, } (t,d,b_2,r_2) \in W$ there holds
$ ((t,d,b_1,r_1), (t,d, b_2, r_2)) \in \check{L}$, i.e., a device
cannot be at different locations at the same time.
Any equivalence relation, like $\check L$, gives rise to a partition
on the base set, like $W$, and vice versa. We can
obtain for $\check L$ a partition of $W$ by
$\check{\mathcal{P}}=W/L=\{[o]_{\check L}\mid o\in W\}$ with the sets
$[o]_{\check L}= \{o' \in W \mid (o, o') \in \check L \}$ called
\emph{equivalence classes}.  Our aim now is to compute a
partition $\mathcal{P}$ of $W$ which approximates $\check{\mathcal{P}}$.
\begin{problem}[Location Identification]\label{prob:locident}
   For a given WiFi data set $W$ with $\check{\mathcal{P}}$, find a partition
   $\mathcal{P}$ of $W$ such that an error function
   $E(\mathcal{P},\check{\mathcal{P}})$ is low.
\end{problem}

Let $\bar L$ be the equivalence relation related to the partition
$\mathcal{P}$, i.e.,
$\bar L = \{(o,o')\in W\times W\mid \exists P\in\mathcal{P}:
\{o,o'\}\in P\}$. The elements of $\mathcal{P}$, representable by
$[o]_{\bar L}$, are called the \emph{observed locations}, as opposed
to the physical locations. Using this we denote by
$L_{d,t} \coloneqq [t,d,b,r]_{\bar L} \in \bigslant{W}{\bar L}$
\emph{the location of device $d$ at time $t$}.
The formulation of \Cref{prob:locident} using partitions reflects our presumed
notion of solving location identification problems through
clustering. This constitutes our ultimate goal. However, in this work
we focus on a simpler variant of this task. By lifting the restriction of
computing a partition in favor of computing a family of sets, we may
employ other approaches to the location identification problem. This
relaxation comes with the price of loosing the strong connection
between partitions and equivalence relations. Hence, we fall back to
an approximation of the true location relation by some relation $L$ on
$W$.

\begin{problem}[Location Discrimination]\label{prob:locdiscrim}
  For a WiFi data set $W$ with $\check L$, compute
  $L\subseteq W\times W$ such that $\hat E(L,\check L)$ is low.
\end{problem}

We will approach~\Cref{prob:locdiscrim} through a distance
measure $d$. For this we suppose we can find for any approximation $L$
some threshold $\tau \in \mathbb{R}_{>0}$ such that for
$O,O' \subseteq W$ with $O\cap O'=\emptyset$ it holds that
$d(O,O') < \tau \iff \forall (o,o') \in O \times O': (o,o') \in
L$. The goal now is to find $L$ through
suitable sets $O\subseteq W$. 

\subsection{Problem Decomposition}
Finding out which WiFi observations belong to the same location in a
data set is a difficult problem due to sensor differences across
devices, measurement noise and signal disturbances through objects,
walls and multi-path effects. We will therefore decompose Problem
\ref{prob:locdiscrim} into two easier problems. First, we restrict our
investigation to the problem where $|D|=1$. Hence, we restrict the
WiFi data set to a particular device d and denote its WiFi data set by
$W_{d}\subseteq W$.
%
Secondly, we will exploit detecting the stationarity of a
device (i.e., a device not in movement). Hence, observations
made during an interval of stationarity can be assigned to one single
location. The recognition of such intervals poses an additional
problem. 
Deciding stationarity or movement (possibly changing locations) will
be done through acceleration sensor data, since acceleration leads to movement and commonly smartphones possess an acceleration sensor. Acceleration typically is measured along three axes, which can be denoted as a vector. We calculate the norm of the vector, because it is invariant to the rotation of a device, therefore not requiring a specific user behaviour of holding the smartphone. 
\begin{definition}[Acceleration data set]
  An \emph{acceleration data set} is a relation
  $A \subset \mathbb{N} \times D \times \mathbb{R}$. We call
  $(t,d,a) \in A$ an \emph{acceleration observation}. The value $a$
  denotes the Euclidean norm of some acceleration vector
  $\mathbf{a} = (a_x, a_y, a_z)^T \in \mathbb{R}^3 $ measured by
  device $d$ at timestamp $t$.
\end{definition}
\begin{problem}[Motion Mode Segmentation]
For a device $d$ in a WiFi data set, find a segmentation of time \(\sigma_{d}=(t_0, t_1, ..., t_n)\) such that $d$ is stationary in the interval $[t_i, t_{i+1})$ if $i$ is even and in movement if $i$ is odd. One such interval is called a segment and two consecutive segments of a device alternate between states of stationarity and movement.
\end{problem} 
From a motion mode segmentation, we infer that  all of the device's WiFi observations made in a stationary segment belong to the same location. 
We will use the segmentation to further restrict $W_d$ to the data where $d$ is stationary, i.e., to the intervals $[t_i, t_{i+1})$ from $\sigma_d$ where $i$ is even. Our disjoint sets of observations, to which we apply our distance measure, will then be constructed by $W_{d,i} \coloneqq \{(t,d,b,r) \in W_d \mid t_{i} \le t < t_{i+1}\}$ for all even $i\in \mathbb{N} < n$. We will call these sets \emph{stationary WiFi segments} or simply \emph{WiFi segments}.

\section{Method}

\usetikzlibrary{shapes,arrows}
\tikzstyle{proc} = [rectangle, draw, minimum height=2.5em]
\tikzstyle{res} = [minimum height=2em, node distance=5cm, text width=5.5em, text centered]
\tikzstyle{line} = [draw, -latex']

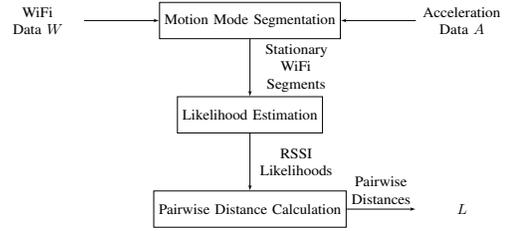
\begin{figure}
\begin{center}
\resizebox{7cm}{3cm}{
\begin{tikzpicture}[node distance = 6.5em, auto, scale=0.10]
\node [proc] (mode) {Motion Mode Segmentation};
\node [res, left of=mode] (wifi) {WiFi Data $W$};
\node [res, right of=mode] (acc) {Acceleration Data $A$};
\node [proc, below of=mode] (dist) {Likelihood Estimation};
\node [proc, below of=dist] (disc) {Pairwise Distance Calculation};
\node [res, right of=disc] (result) {$L$};
\path [line] (acc) -- (mode);
\path [line] (wifi) -- (mode);
\path [line] (mode) -- node[res]{Stationary WiFi Segments}  (dist);
\path [line] (dist) -- node[res]{RSSI Likelihoods} (disc);
\path [line] (disc) -- node[res]{Pairwise Distances}(result);
\end{tikzpicture}
}
\end{center}
\caption{Visualization of our approach. Blocks indicate processes. Arrows are annotated by in-/outputs.}
\label{fig:method}
\end{figure}
Our method, as depicted in Figure \ref{fig:method}, works as follows:
First, we perform motion mode segmentation based on the acceleration
data. We then estimate the RSSI likelihood for each
stationary WiFi segment. In particular we investigate on various methods to estimate
the likelihood based on discrete and continuous probability
distributions. To quantify dissimilarities between the likelihoods, we
calculate their pairwise distances through a distance measure on
probability distributions. We omitted the step \emph{location discrimination}
here, which we will present in~\Cref{sec:experiments}.

\subsection{Motion Mode Segmentation}
Motion mode segmentation is done in a bottom-up-approach consisting
of two parts. First, we represent the acceleration data of a device as
a time series $\{a_t\}_{t\in \mathbb{N}}$ and classify short time
intervals as either movement or stationarity. We do this by applying a
sliding window to the series, calculating a window function and
thresholding it for classification. As a window function, we used the
energy \cite{motion_survey}. However, utilizing the sample variance
gave similar results. We determined a decision threshold through a
decision tree. This method achieved about 96\% accuracy in a 10-fold crossvalidation on a small activity recognition data set. 
Secondly, we aggregate consecutive windows from the same class. This
gives us the final \emph{motion mode segmentation}, i.e., an
$n$-tuple of the timestamps where modes of movement start and end. 

\subsection{Segment Representation}\label{sec:segment_representation}

We assume that the true likelihoods are unique and thus we can
recognize and distinguish locations through the dissimilarities
between likelihood estimates made from WiFi observations. Formally, we
assume there is a conditional probability distribution
$p(r_{1},r_{2}, ..., r_{m}\,|\, L_{d,t})$ for each location $L_{d,t}$
where $r_{k}$ is a random variable for the RSSI measured from access
point $b_k$. This conditional distribution is the presumed true
likelihood of observing a combination of RSSI values given the
location. We will also make the assumption that given a location, the
signal strenghts of different access points are conditionally
independent. Thus, their conditional joint distribution can be
factorized.
\begin{equation}\label{eq:pmf}
p(r_{1},r_{2}, ..., r_{m}\,|\, L_{d,t}) = \prod_{k=0}^{m}{p(r_{k} \,|\, L_{d,t}})
\end{equation}
This assumption has been frequently made for representing RSSI
likelihoods in previous research (e.g., in \cite{horus2005}). However,
there are some arguments against it. For example, different rotations
of the person carrying a device cause signals from varying directions
being damped by the body. Clearly, if a different rotation is not
considered as a different location, then there is some dependence
between signals. Furthermore, we assume implicitly that the
distribution is independent of time, while in reality this clearly is
not the case. As an example, a change in the environment can change
the signal distribution. However, we assume that our approximation in
combination with a distance measure is still close enough to
rerecognize locations in reasonably short time spans and robust enough for small changes in the environment or rotations of the body.

We consider the WiFi observations made during a stationary segment as samples from the underlying distribution. Thus, for each stationary WiFi segment $W_{d,i} \subseteq W_d$ with samples from the interval $[t_i, t_{i+1})$, we calculate an estimate $\hat{p_i}$ of the underlying true likelihood $p_i$. 

In \Cref{eq:pmf} we assumed that the conditional joint distribution of the RSSI values is a product of the RSSI likelihoods of each single AP. In the following, we will therefore introduce several methods for estimating the factors in \eqref{eq:pmf}, i.e., the RSSI likelihoods for single access points.

\subsubsection*{Representation through a Probability Mass Function}
We frst consider estimating the RSSI likelihood of an access point as a discrete probability mass function (PMF). Hence, we model each factor in \eqref{eq:pmf} as a normalized histogram with bin size one, i.e., to each possible RSSI-value we assign the probability mass of its relative frequency. With our WiFi data set this can easily be done through counting the RSSI values, which are given as integers.

Formally, let $W_{d,i}^{k} \coloneqq \{(t,d,b_k,r) \in W_{d,i}\}$ be a device's measurements of the signal strengths received from access point $b_k$ during the segment $[t_i, t_{i+1})$. 
We estimate the probability of receiving a signal strength of $r'$ from access point $b_k$ given the location $L_{d,t}$ of the segment $[t_i, t_{i+1})$ as follows:
\begin{equation}\label{eq:pmfestimate}
\hat{p_i}(r_{k}=r' \,|\, L_{d,t}) \coloneqq \frac{|\{(t,d,b_k,r') \in W_{d,i}^{k}\}| }{|W_{d,i}^{k}|}
\end{equation}
If $|W_{d,i}^{k}|=0$, i.e., no observations are made for $b_k$ during
the segment, we assign the full probabillity mass of 1 to an RSSI of
$-100$. This is slightly lower than the lowest value we observed in
any experimental measurements. Thus we model the situation of an
invisible access point as observing a very low signal strength. This
ensures that we have valid probability distributions for all APs, even
when no values were observed. Also this representation is useful for
calculating distances. If an AP is visible in one WiFi segment and
invisible in another this will increase the distance between their
RSSI likelihoods.

\subsubsection*{Representation through a Normal Distribution}
This estimation is based on the assumption that the true RSSI likelihoods are normally distributed. We thus estimate $\hat{p_i}$ by
\begin{equation}
\hat{p}_i (r_k=r' \mid L_{d,t}) \sim \mathcal {N} (\hat{\mu}_i, \hat{\sigma}_i^2)
\end{equation}
where $\hat{\mu}_i$ and $\hat{\sigma}_d^2$ are the sample mean and sample variance of the RSSI values of one AP in a WiFi segment. Formally, let $R_{d,i}^k = \{r' \mid (t,d,b_k,r') \in W_{d,i}^k\}$ be the RSSI values of AP $b_k$ in a WiFi segment. Then we calculate $\hat{\mu}_i$ and $\hat{\sigma}_i^2$ for $b_k$ by:
\begin{equation}
\hat{\mu}_i = \frac{1}{| R_{d,i}^k |} \sum_{r\in R_{d,i}^k}{r},\quad
\hat{\sigma}_i^2=\frac{1}{| R_{d,i}^k | -1} \sum_{r\in R_{d,i}^k}{(r-\hat{\mu}_i)^2}
\end{equation}


\subsubsection*{Representation through Kernel Density Estimation}
Kernel density estimation (KDE) is a technique for estimating a continuous probability density function (PDF) from a given set of samples from the underlying distribution. We will explain KDE for the univariate case here.
Let $\{x_1, x_2, \cdots, x_n \}$ be a set of samples from a univariate, continuous random variable $X$. Then the kernel density estimate $\hat{p}$ of the PDF of $X$ is
\begin{equation}
\hat{p}(x) = \frac{1}{nh} \sum_{i=1}^{n}{k\left(\frac{x-x_i}{h}\right)},
\end{equation}
where $k$ is called a \emph{kernel} or \emph{kernel function} and subject to
$k(x) \ge 0$ and $\int_{-\infty}^{\infty} {k(x)dx}=1$. The parameter
$h\in \mathbb{R}_{>0}$ is the \emph{bandwidth} of the kernel. A higher
bandwith leads to a smoother probability density estimate. In our
experiments, we use the Gaussian kernel, which leads to the probability density estimate being a mixture of $n$ Gaussians with standard deviation $h$ and means located at the sample locations $x_i$:
\begin{equation}\label{eq:gausskernel}
k(x) \coloneqq \frac{1}{\sqrt{2\pi}} \exp { \left( -\frac{1}{2} x^2 \right)}
\end{equation}
\subsubsection*{Laplace Smoothing}
Some of the considered distance measures, which we will introduce in
the next section, cannot handle zero probabilities. As an example, the
symmetrized Kullback–Leibler (KL) divergence is undefined when $p(x)$
or $q(x)$ is zero for an outcome of a random variable $x$. The
Bhattacharyya distance is undefined when $BC(p,q)=0$, i.e., when the
distributions $p$ and $q$ do not overlap. This is because the
logarithm of zero is undefined and goes to negative infinity as $x$ goes
to $0$.
This problem does not occur when RSSI likelihoods are estimated
through a normal distribution or through KDE with a Gaussian kernel,
since all densities are then strictly positive. However, when a
discrete PMF is employed, we apply Laplace smoothing to compensate for
this problem, i.e., we add a small constant to each probability and
normalize the probabilities such that their sum is one. This procedure is justifiable through Cromwell's rule \cite{decisiontheory}, which states that one should never assume a zero probabilitiy of an outcome of a random variable, if one cannot be absolutely sure that it will never occur. Certainly, we cannot be absolutely sure that an RSSI value can not occur at a location, just because we did not observe it in our measurements. In fact it is very likely that we will observe some different RSSI values, especially when we estimated the likelihoods from few samples. Zero probabilities also could easily break further inferences, e.g. location estimation through maximum likelihood, since the product of the likelihoods of an observed RSSI vector would become zero, when the likelihood of an RSSI value is zero at a location just for one single AP.
\subsubsection*{Modelling Probabilities of AP Invisibilities}
For a (smartphone) device scanning all WiFi channels in the 2.4 GHz
and 5GHz band takes about three seconds in practice. Hence, only one
RSSI value per AP can be measured during such a scan. We may therefore
obtain sometimes too  few samples in cases where (stationary) segments are short.
%
%
Also, in practice, many of the RSSI values will be overlooked by the
(smartphone) device during a scan, i.e., the AP will be invisible to
the scanning device. The probability of observing or missing an RSSI value is a distinct feature of a location by itself and therefore provides further useful information for characterizing locations. Approximately, the probability of AP invisibility is decreasing with increasing signal strength of the AP. 
We model this probability by extending our WiFi data set through
\textsl{pseudo-observations} of $-100$ RSSI values for all APs that were
not observed at a recorded timestamp. Formally, for every distinct
pair of timestamp and device $(t', d')$ s.t. $\exists (t', d', b, r)
\in W$ and for every AP $b' \in B$, if there exists no $(t',d',b', r)
\in W$ we add an observation $(t', d', b', -100)$ to $W$. Then we
proceed as before to estimate likelihoods, where the probabilitiy of a
$-100$ RSSI now models the probability of AP invisibility at a location, even when some RSSI values were observed from it.


\subsubsection*{Comparison}
There are several advantages and disadvantages of the presented
likelihood representations, which we would like to point here.  An
advantage of using both the normal distribution and the kernel density
estimation is that similar probabilities are also assigned to values
close to the observed RSSI values. Often it can be the case that a
specific RSSI value does not occur in a sample by chance. If only
relative frequencies are considered, this leads to a zero probability
estimate. However, it is often more appropriate to assume that
outcomes close to the observed values can occur with similar
probability. As an example, take the case where the RSSI values -70
and -68 have been observed several times during a segment but never
the value -69.

A disadvantage of representing segments through a continuous PDF is
that assumptions have to be made about the distributions. If a certain
distribution is fit to the data, one has to make an assumption about
what the underlying distribution of the data is. For RSSI likelihoods,
the normality assumption has often been made in literature
\cite{mirowski2}. In experiments related to the presented
in~\cref{sec:experiments} we experienced cases where a smartphone
lying around at the same position often had regular down peaks for a
strong AP. This means the RSSI was nearly constant for longer
intervals and would once in a while decrease by approximately 10dBm
for short intervals. The resulting distribution of the measurements
therefore has two modi and is not normally
distributed.

If kernel density estimation is employed, arbitrary distributions can
be approximated \cite{Bishop}. However, one has to select a bandwidth
parameter, which determines how much probability is assigned to the
outcomes close to the observations. Choosing a suitable bandwidth
parameter is then a task by itself. Another disadvantage of a
continuous PDF is that calculations can become computationally more
expensive, since summation becomes integration. To compute the density
at some point we need to evaluate the kernel function for every
distinct sample, whenever kernel density estimation is used.
Therefore KDE has additional computational costs.

\subsection{Distance Calculation Between Segments}
There exists a wide variety of distance measures between probability
distributions. We denote the PDF or PMF of a distribution through
small letters $p$ and $q$ and the cumulative distribution functions
(CDF) through big letters $P$ and $Q$. We then consider the distance
measures listed in \Cref{tab:dist_measures}.  \bgroup
\def\arraystretch{1.5} 

\begin{table*}
  \caption{Distance measures between probability distributions and
    their properties. Note that the Bhattacharyya coefficient is not a
    distance but a similarity measure, but is crucial for the
    definitions of the Hellinger distance and the Bhattacharyya
    distance. We included KL divergence for a similar reason.  By Id.\
    of indiscernibles we denote the property
    $d(x,y)=0 \iff x=y$.}
\begin{center}
\begin{tabular}{| l | l | c c c |}
\hline
Name & Equation & Id. of Indiscernibles & Symmetry & Triangle Inequality\\
\hline
KL Divergence~\cite{ref_kl} & $KL(p\parallel q) = \int{p(x)\log\frac{p(x)}{q(x)}
                \mathrm dx}$ & Y & N & N \\
Symmetrized KL Divergence~\cite{ref_kl} & $D(p,q) = KL(p\parallel q)+KL(q\parallel p)$ & Y & Y & N \\
Jenssen Shannon Divergence~\cite{lin91divergence} & $JSD (p, q)={\frac{1}{2}  (KL(p\parallel m)+KL(q\parallel m)) }\text{, } m\coloneqq\frac{p+q}{2} $ & Y & Y & N \\
\hline
Bhattacharyya Coefficient \cite{ref_bhatt}& $BC(p,q)=\int{\sqrt
                                            {p(x)q(x)}\mathrm dx}$ & - & - & -\\
Bhattacharyya Distance~\cite{Basseville1989} & $D_{B}(p,q)=-\ln \left(BC(p,q)\right)$ & Y & Y & N \\
Hellinger Distance~\cite{Basseville1989} & $H(p,q)={\sqrt {1-BC(p,q)}}$ & Y & Y & Y \\
\hline
Kolmogorov-Smirnov Distance~\cite{Basseville1989} & $D(p,q)=\sup _{x}|P(x)-Q(x)|$ & Y & Y & Y\\
Earth Mover's Distance~\cite{ref_emd2} & $EMD(P,Q)=\int{|P(x)-Q(x)|\mathrm dx}$ & Y & Y & Y\\
\hline
Absolute Difference of Means & $D(p,q)= \left| \mathbb{E}_p[x] -
                               \mathbb{E}_q[x] \right|,\
                               \mathbb{E}_p[x] = \int{p(x)x\,\mathrm dx }  $ & N & Y & Y\\
\hline
\end{tabular}
\end{center}
\label{tab:dist_measures}
\end{table*}
\bgroup
\def\arraystretch{1} 


Since our full joint distributions are high-dimensional (i.e.,
multivariate), computations for many of these distance measures become
intractable due to the increasing computational complexity of the
quadrature. Others, e.g., the Kolmogorov-Smirnov distance (which is
actually a test statistic), cannot easily be extended to the
multivariate case. Therefore we calculate distances between the
univariate RSSI likelihoods of single APs and take the sum over all
APs (i.e., also those that are invisible).  Since all distances are
positive, this is the same as calculating the $L_1$-norm.
In case of the KL divergence it can be shown that if the
univariate distributions are independent, this
sum 
is the KL
divergence of the joint distributions \cite{ref_kl}.
To our knowledge the same does not hold for the other distance
measures with exception of the symmetrized KL divergence. However, we
think that it is reasonable to increase the total distance
proportionally with every AP distance. For comparison we also
calculate the $L_2$-norm, which in turn puts more weight on larger
distances. Our distance measure between two multivariate signal
strength distributions $p$ and $q$ is therefore defined as
\begin{equation}\label{eq:dist_measure}
d_\ell(p, q) = \Big( \sum_{k=1}^m {d(p(r_k), q(r_k))^\ell} \Big)^{1/\ell},\quad\ell \in \{1,2\},
\end{equation}
where $p(r_k)$ and $q(r_k)$ are the univariate signal strength
distributions of AP $b_k$ and $d$ is a distance measure from~\Cref{tab:dist_measures}.


At this point we would like to note that, if the used distance measure
between single APs is a bounded metric, then also our distance measure
$d_\ell$ is a metric. Also note that APs invisible in both segments,
i.e., both compared distributions, can be omitted to
compute~\Cref{eq:dist_measure}. This follows from~\Cref{def:distx}.

Considering the distance measures in Table \ref{tab:dist_measures}, we
would like to point out that calculating the absolute difference
between the expected values is not a metric on probability
distributions, since different distributions can have the same
mean. Nonetheless, it is a metric on the expected values. Taking the absolute difference of means makes our distance measure equivalent to computing the Manhattan or Euclidean distance between the average RSSI vectors of two segments.


Other work \cite{ref_kl} suggests using the symmetrized KL divergence for supervised localization through kernel regression. 
However, this measure has two disadvantages for the comparison of RSSI likelihoods: First of all, it is not a metric, because the triangle inequality is not required. Distances between physical locations, viewed as points in $\mathbb{R}^3$, follow the triangle inequality. Therefore using a metric to compare location representations is more consistent with our view of the physical world. 
Secondly, consider the case where two distributions do not overlap and
there is a gap between them. 
Then symmetrized KL divergence reaches its maximum  value no matter
how large the gap is. However, it makes no difference how large the gap between the distributions is. The Earth Mover's Distance (EMD, \cite{ref_emd2}) on the other hand becomes larger, the larger the gap between the distributions is. In the case of RSSI likelihoods, it makes sense to infer that locations are farther away from each other when the gap between RSSI likelihoods becomes larger.

\section{Experiments}
\label{sec:experiments}
To test our method, we conducted an experiment at our group located in
Kassel, Germany. We collected data for five days during work
time. Each participant was instructed to carry a smartphone with her
as well as an RFID badge clipped to the
chest~\cite{cattuto2010dynamics}. We recorded WiFi and acceleration
data on each smartphone using the \emph{Sensor Data Collection
  Framework}~\cite{ref_sdcf} (SDCF) for the Android operating
system. The RFIDs were used to collect ground truth data. For this we
installed stationary RFIDs at several physical locations we considered
important. These locations are the desks of participants, which were
located in different rooms, our coffee kitchen and a table soccer. The
idea here is that whenever two persons meet we can register a
\emph{contact} using the RFID badges. This encounter would be recorded
and sent to our servers. The same is true for an encounter of a person
with the afore mentioned desks, kitchen or tables soccer. More
technically, a contact occurs, when two RFIDs are within a distance of
approximately 1.5 meters. Since human bodies block the signal, a
contact only occurs when a person's chest points roughly towards
another RFID. Participants were additionally instructed to fill in
manual logs about their locations and activities, which can be used
for plausibility tests. In total, eight people took part in the
experiment. To ensure device heterogenity, we used several different
smartphone models, as depicted in \Cref{tab1}. One of the participants
carried two
smartphones.

\begin{table}
\caption{Experiment participants and used smartphones.}
\centering
\begin{tabular}{lcrr}
\hline
PersonID & Device Model & 5GHz & Data used\\
\hline
\hline
1 &Samsung GT-I9195&  Y & N\\ 
2& LG Nexus 4& Y & Y\\
3 & LG Nexus 5  & Y & Y\\
3 & HTC One X+ & Y & Y\\
4 & OnePlus 5 & Y & N\\
5 & HTC One X+ & Y & Y\\
6 & Jiayu S3+ & Y & N \\
7 & Samsung GT-I9001& N & N\\  
\hline
\end{tabular}\label{tab1}
\end{table}
Some smartphone models could not reliably record data over longer
periods of time. Some minutes after starting the data collection
process through the SDCF the collection process was unintentionally
stopped. We suspect this problem occurs due to vendor specific battery
saving measures integrated into the operating system. For other
participants, we could not collect much data because they were mostly
not present at our group during the experiment or the batteries of
their devices did not last long enough. In total, we used the data
from four devices for our experiments. Altogether, we collected about
1.38 Million WiFi observations adding up to roughly 247 hours. The
data from the four smartphones used in the evaluation adds up to about 40
hours on average per device.

\subsection{WiFi Data Preprocessing}
In some cases, RSSIs from mobile APs, e.g., hotspots raised by mobile
phones themselves, are observed. Since these APs are likely to change
their location, we ignore observations from them, i.e., remove them
from the WiFi data set.  These APs are often recognizable through
their Service Set Identifier (SSID), i.e., the name of the WiFi
network. Examples of such APs are \textsl{Tim's iPhone},
\textsl{AndroidHotspot} or \textsl{Porsche}. Due to a parking lot next to
our building, many of such car hotspots appeared in our
recordings. We remove mobile APs through a manually crafted blacklist
of SSID-prefixes and postfixes. However, we observed that the relative
amount of such recordings was comparably low. Hence, we think their
negative influence on localization is not very strong and often
mitigated through other APs, even if some mobile hotspots are
overlooked. 
As a final preprocessing step, we only include stationary WiFi segments with a minimum duration of ten seconds in our analysis. This is because we think that shorter durations contain too few samples to estimate RSSI likelihoods.

\subsection{Ground Truth Preprocessing}

\begin{figure*}
\centering
\includegraphics[width=0.3\textwidth] {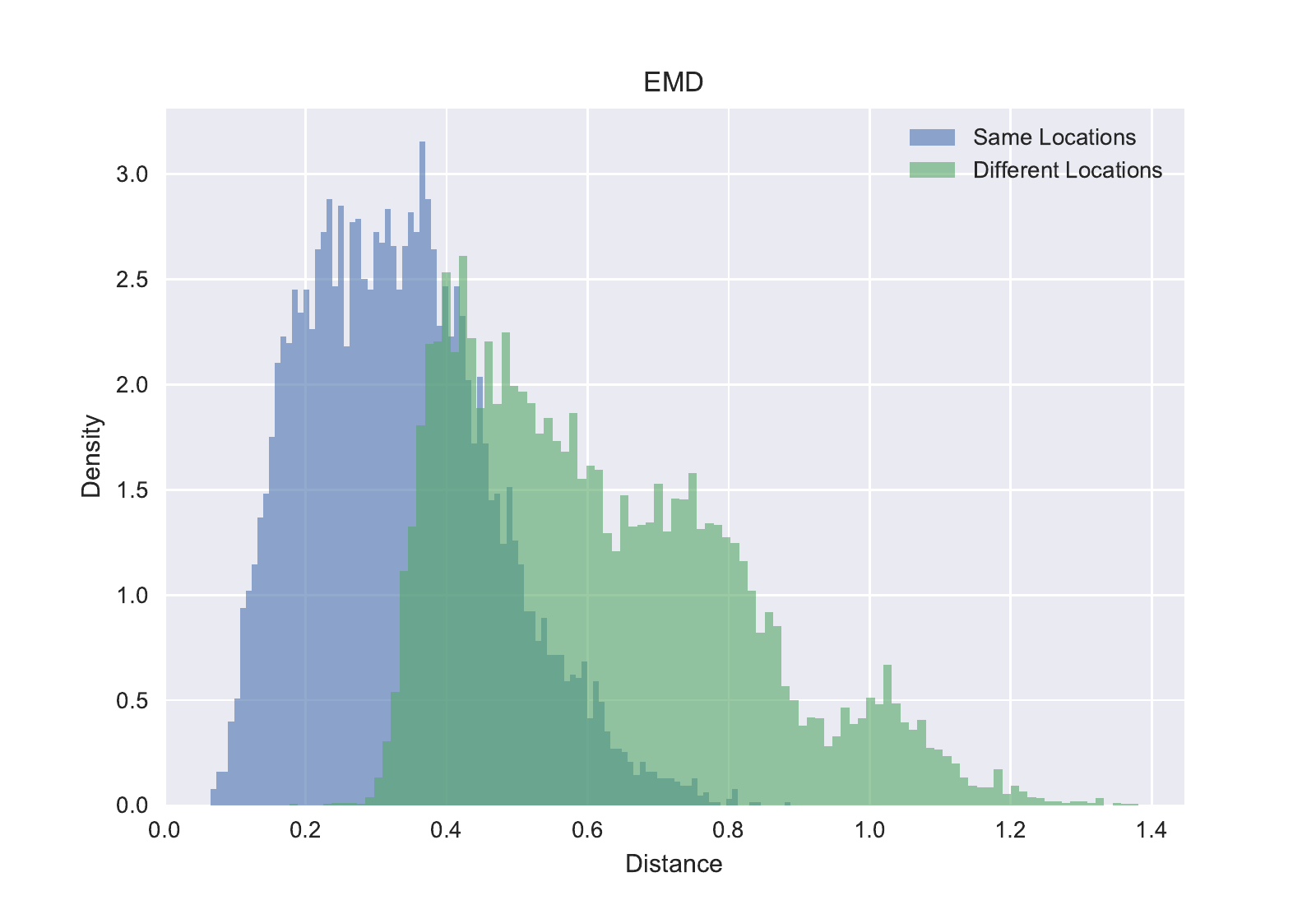}
\includegraphics[width=0.3\textwidth] {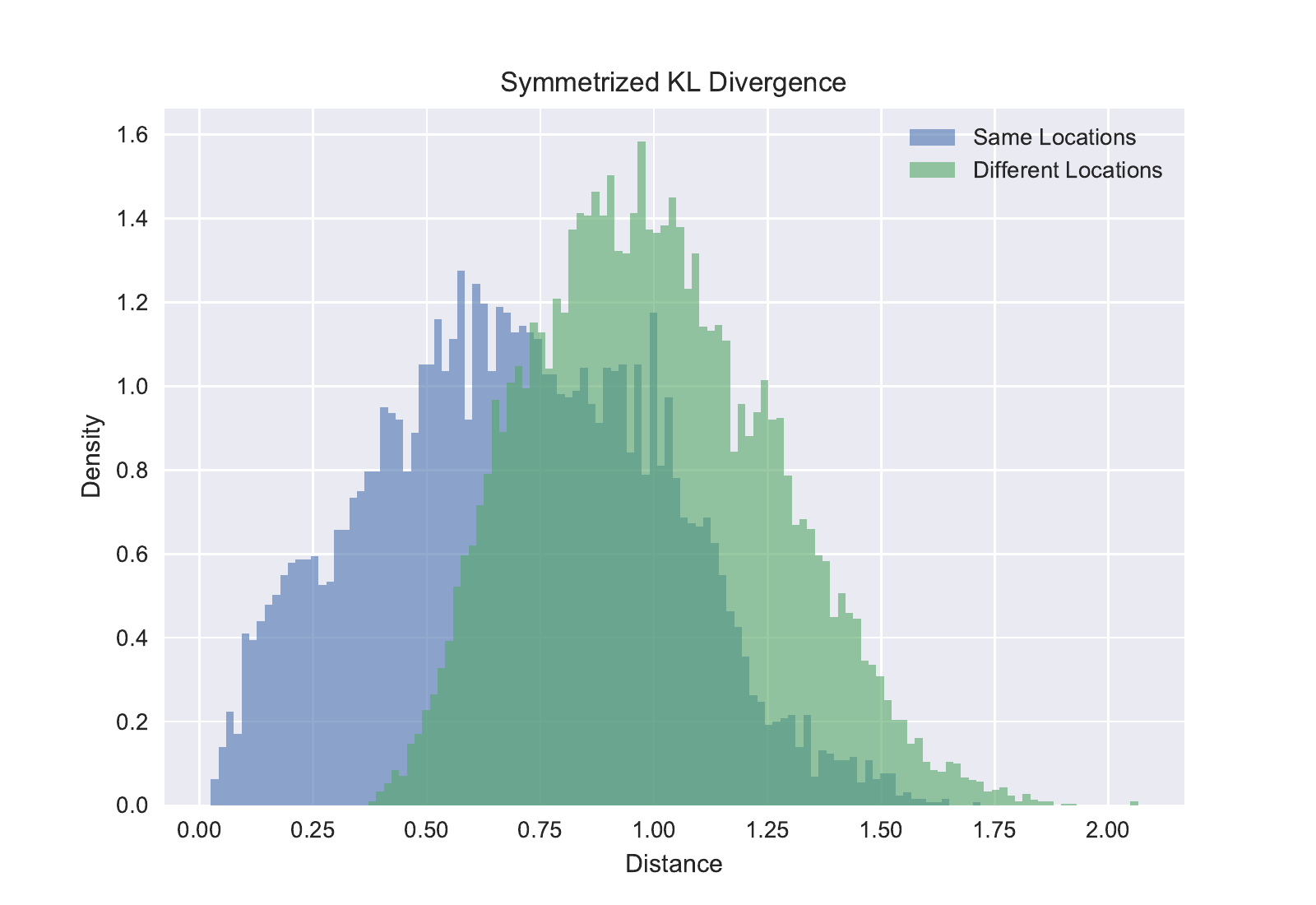}
\includegraphics[width=0.3\textwidth] {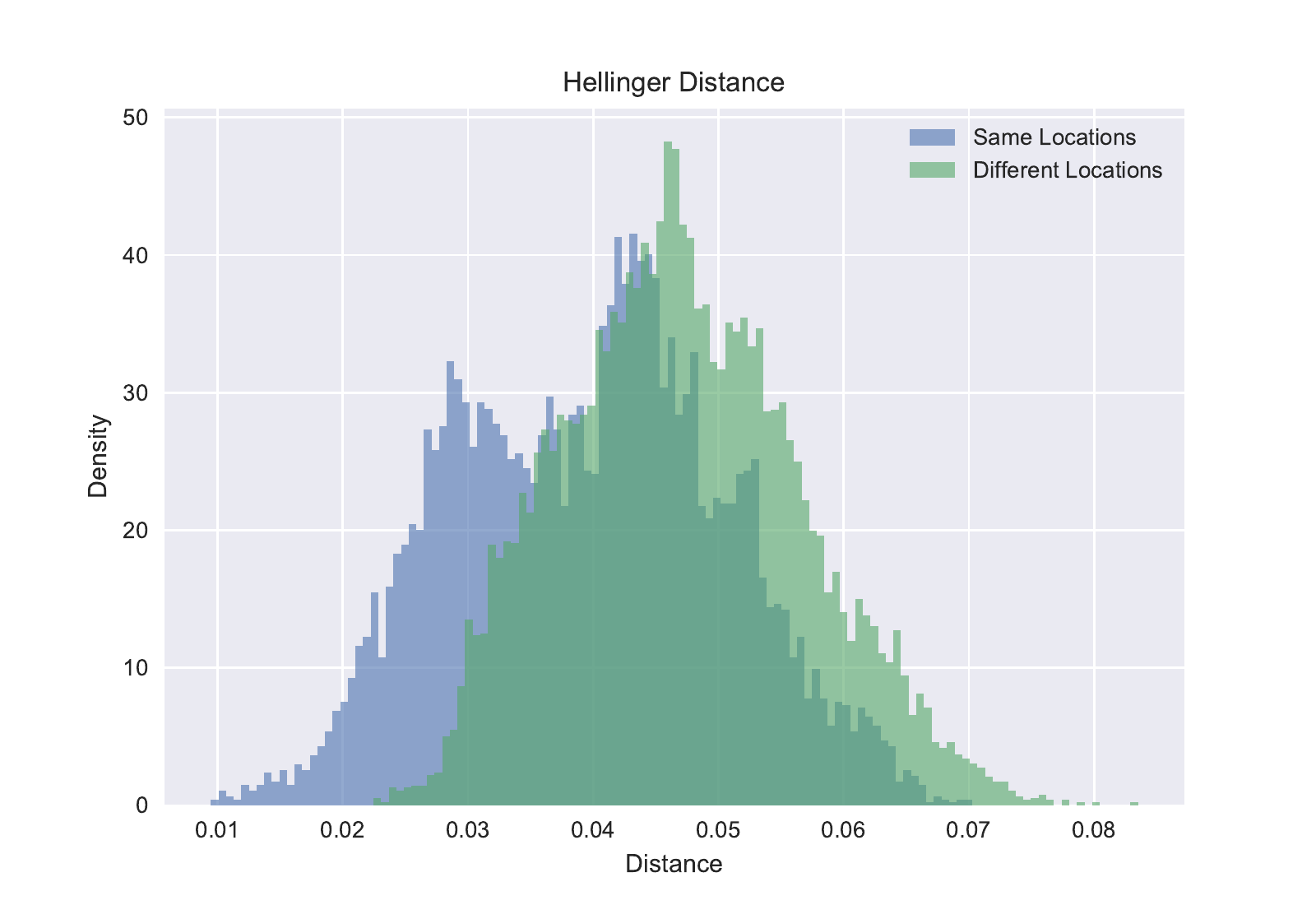}
\caption{Normalized histograms of calculated distances for LG Nexus 4. In each plot, the left histogram depicts distances between segments from the same location, the right histograms distances between different locations. Likelihoods were estimated through PMF and invisible APs included.} \label{fig:dist_histograms}
\end{figure*}

Ground truth data from RFIDs is sparse. As addressed before, a
person's spatial orientation can lead to a (human) body blocking the
RFID signal. Additionally, objects interfering with the 2.4GHz or 5GHz
band do block or damp signals. We therefore enhance our ground truth
in two ways: First, we make use of the symmetry of contacts. If an
RFID badge $a$ receives a contact signal sent from an RFID badge $b$,
we enhance our data by adding a pseudo contact signal received by $b$
from $a$. Secondly, we aggregate contacts, which were initially
recorded at distinct timestamps, to time intervals. We assume that for
two fixed badges, if the timestamp difference between two contacts is
lower than a given threshold, then there has been a contact for the
whole interval duration. As a threshold we use the difference of one
minute. We chose this value based on the comparison of RFID data to
the manual logs of the participants. However, there is a trade-off
between correctness and availability of ground truth. While a too big
interval threshold can introduce errors to the ground truth, a too
small one leads to sparser data.

\subsection{Discriminative Evaluation}

\begin{figure}
\centering
\includegraphics[width=0.24\textwidth] {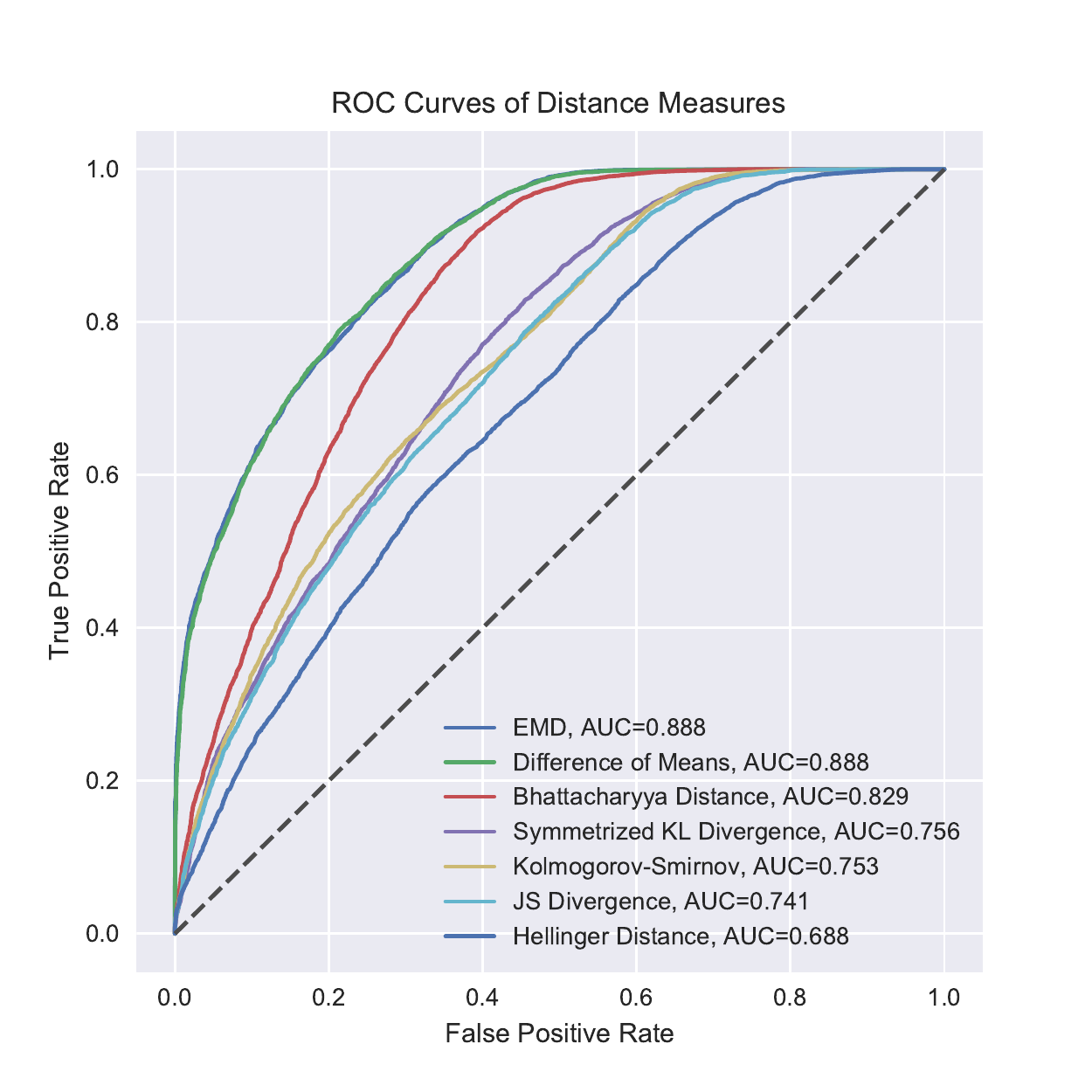}
\includegraphics[width=0.24\textwidth] {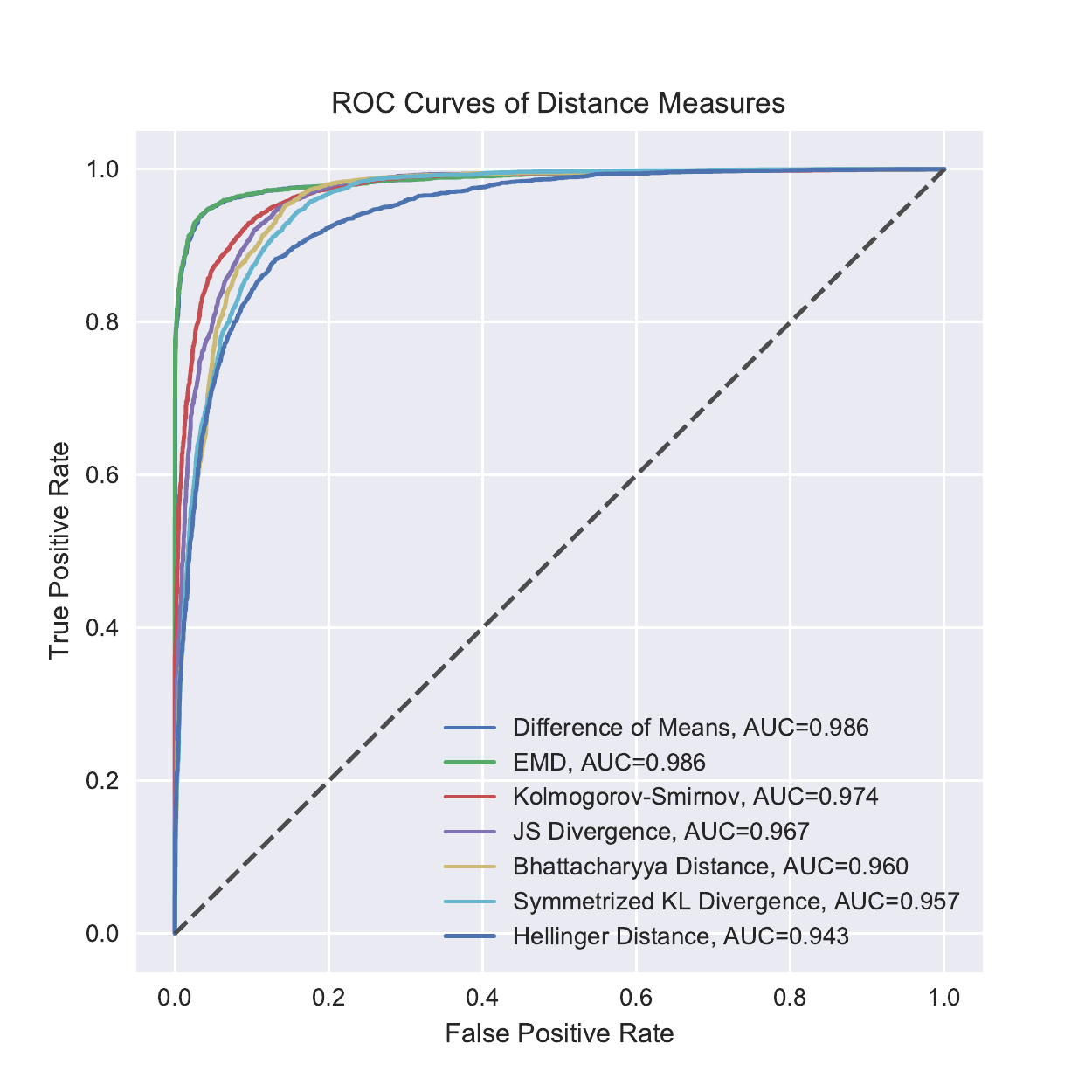}
\caption{ROC Curves: HTC One X+ (left), Nexus 5 (right). For all
  distance measures, likelihoods were estimated as PMF and AP invisibilities included.} \label{fig:roc1}
\end{figure}

One aim of using a distance measure on RSSI likelihoods is to find out
whether two sets of observations were made at the same physical
location. This task can be treated as a binary classification
problem. 
Based on this we developed the following evaluation scheme.  We
represent each set of observations made at one day in one room as a
signal strength distribution as introduced in
\Cref{sec:segment_representation}. Using this we then calculate the
distances between all possible pairs of signal distributions from the
same room and from different rooms. \Cref{fig:dist_histograms} shows
histograms of the calculated distance values. The left histogram there
in each plot contains the distances calculated between the same room.
The histograms to the right there in each plot depict distances
calculated between different rooms. The lower the overlap between the
histograms, the more capable our distance measure is of
discriminating rooms.

Let us now consider a binary classifier which discriminates the same
from different locations by thresholding (see~\Cref{sec:formalProb})
on the distance measure~\Cref{eq:dist_measure}.

\begin{equation}
f: P\times P \rightarrow \{0,1\} \text{, }  f(\hat{p}_i, \hat{p}_j) = \Theta(d(\hat{p}_i, \hat{p}_j)-\tau)
\end{equation}
In this equation $\Theta$ is the Heaviside step function which returns
zero if its argument is smaller than zero, and one otherwise. The
value $\tau$ is a threshold. Hence, our classifier returns one if the
distance between the two input likelihoods is not smaller than $\tau$.
Semantically one is interpreted as the likelihoods stemming from
different locations. On different levels of $\tau$, we calculate the
\emph{true positive rate} by $tpr\coloneqq tp/(tp+fn)$ and \emph{false
  positive rate} by $fpr\coloneqq fp/(fp+tn)$ of this classification
function on our WiFi data set. In this $tp$ and $fp$ denote the number
of true positives and false positives, and $tn$ and $fn$ the number of
true negatives and false negatives.  We then calculate the area under
the ROC curve (AUC) to evaluate our distance measure. AUC can be
interpreted as the probability of ranking a random positive sample
higher than a random negative sample \cite{roc}. In our case this is
the probability of assigning a higher distance to a random pair of
segments from different locations than to a pair of segments
from the same location. An example of the resulting ROC curves and AUC
values is given in Figure \ref{fig:roc1}.

\subsection{Evaluation of Correlations}
\label{sec:evalcor}
\begin{figure}
\includegraphics[width=0.48\textwidth] {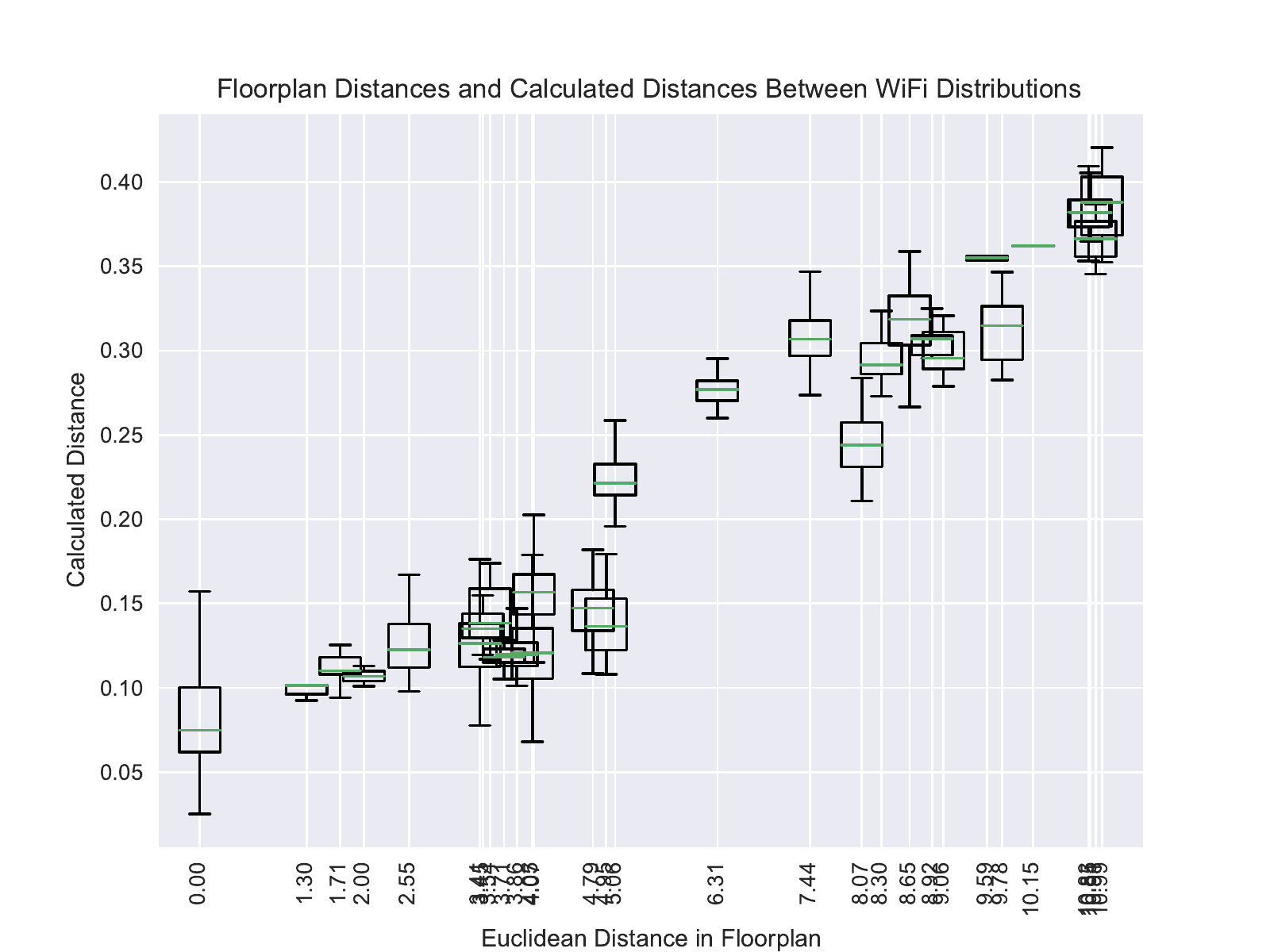}
\caption{Correlations between floor plan distances
  and calculated distances between RSSI likelihoods
  for one device. Left: Several distances are calculated for the
  same pairs of true locations, hence distributions of the calculated
  distances are given through box plots. Right: We present the mean
  calculated distance for the y-coordinate.  Euclidean distance
  between distribution means were applied. Correlations: Pearson 0.94, Kendall: 0.79, Spearman; 0.92.}\label{fig:dist_correlations}
\end{figure}

To evaluate how well the considered distance measures capture
distances between physical locations, we proceed as follows: For the
ground truth locations, we determine the position of tags in a floor
plan and measure their pairwise distances. We calculate the Pearson
correlation coefficient, Spearman's rank correlation coefficient and
Kendall's tau between calculated and floor plan distances. The Pearson
correlation coefficient measures the degree of linear relationship
between two variables. Our ratio is that a perfect distance measure
would provide perfect linear correlation with floor plan
distances. Spearman's rank correlation coefficient measures the
monotonic relationship between variables. In our case this means how
well the ranking of the calculated distances matches the ranking of
the floor plan distances. Kendall's tau is a different way to
calculate the rank correlation and thus expresses a similar measure as
the Spearman rank correlation. Our example
in~\Cref{fig:dist_correlations} shows a result from our experiment
where these correlations are captured for a particular distance
measure. In this case a strong correlation can be observed.



\subsection{Visual Evaluation}
We use multidimensional scaling (MDS) to layout WiFi segments in
$\mathbb{R}^2$. This method receives a pairwise distance matrix as
input and determines coordinates in $\mathbb{R}^{n}$ for a given $n$
such that the distances are preserved optimally. Our ratio here is
that the result retains at least topological relations between
locations, i.e., segments from the same location should be kept close
together. We show an exemplary result of this procedure
in~\Cref{fig:mds}, where we use the Manhattan distance between
expected values. The results for other participants are similar.

\begin{figure}
\includegraphics[trim=0 40 0  40, clip,width=\columnwidth]{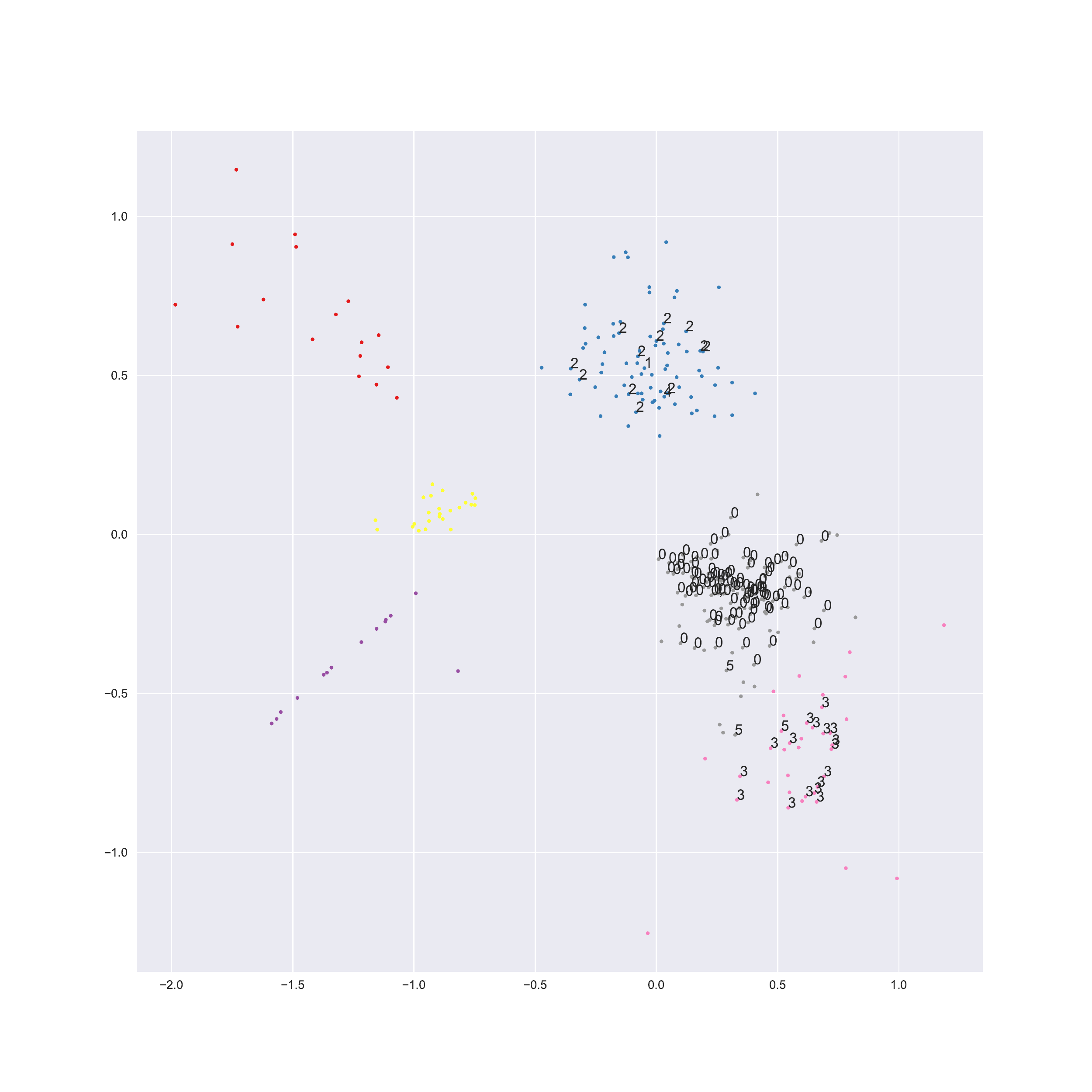}
\caption{Plot of the calculated MDS coordinates for one smartphone. Each point represents a WiFi segment. The segments have been clustered through the HDBSCAN algorithm \cite{ref_hdbscan} and colored according to their cluster. Points are annotated by RFID contacts, which give hints about true locations.}\label{fig:mds}
\end{figure}
The results indicate, that indeed topological structures are retained
through our method. We recognize six main clusters, not of all which
are annotated through ground truth data. The ground truth is as
follows: 0-desk, 5-desk in neighbored room, 3-table soccer, 2-kitchen,
1-desk near the kitchen and 4-table directly next to the kitchen. The
other three clusters could be identified from our manual logs as a
bathroom, the canteen and a food store located next to our
building. For the annotated locations, topological relations were
mostly retained, i.e., the bureau labeled by 0 is indeed halfway in
between the other two locations. However this does not hold for the
relations between the three unlabeled locations. These locations are
very far away from each other and do not have any APs in common. There
are specific areas, where locations are hardly distinguishable through
WiFi signals alone. We attribute this to little AP coverage and thin
walls, since we also found these difficulties in our related supervised
localization
experiments. 



\subsection{Discussion}
We calculated ROC curves and correlations for the four used
smartphones and all distance measures in all kinds of combinations of
modelling likelihoods and including AP invisibilities in the
distributions or not. In total, we consider seven distance measures,
each with $L_1$ norm and $L_2$ norm. We have three ways of modelling
distributions and two possibilities for counting invisible APs or not
counting them. Altogether we have $84$ combinations of calculating
distances for each smartphone and therefore $336\cdot4$ performance
measures to compare. Due to limited space we may only sum up our main
findings. The full evaluation results as well as our WiFi data set are available
from our web page (\url{https://kde.cs.uni-kassel.de/datasets}). 

In the results we observed in general, that measures with
better AUC implied better correlations with floorplan
distances. Similarly, when one correlation was stronger, the other
correlations were stronger. We addressed this partially
in~\Cref{sec:evalcor}.

\subsubsection{Distance Measures}
\begin{figure}
\begin{centering}
\includegraphics[width=0.48\columnwidth]{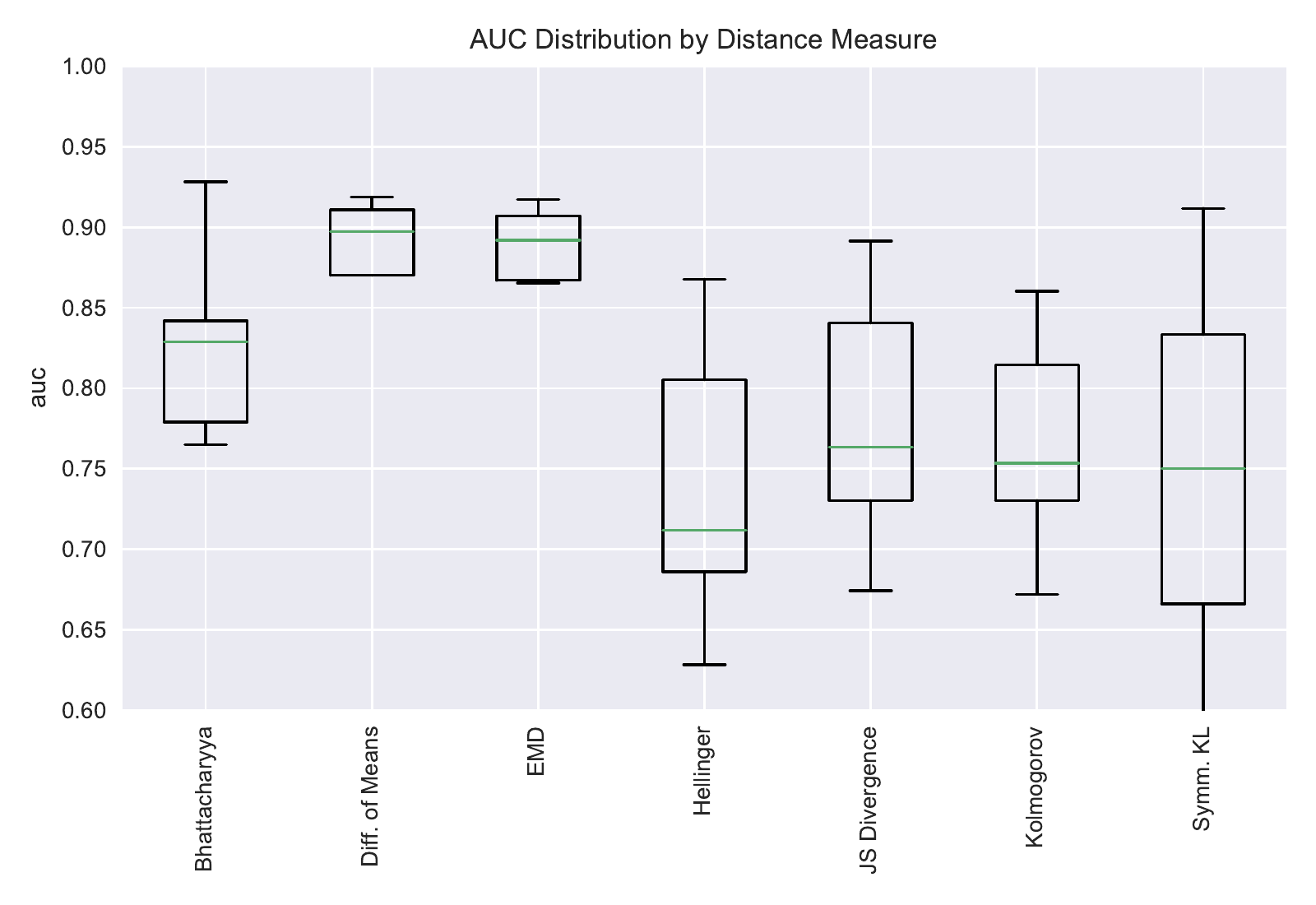}
\includegraphics[width=0.48\columnwidth]{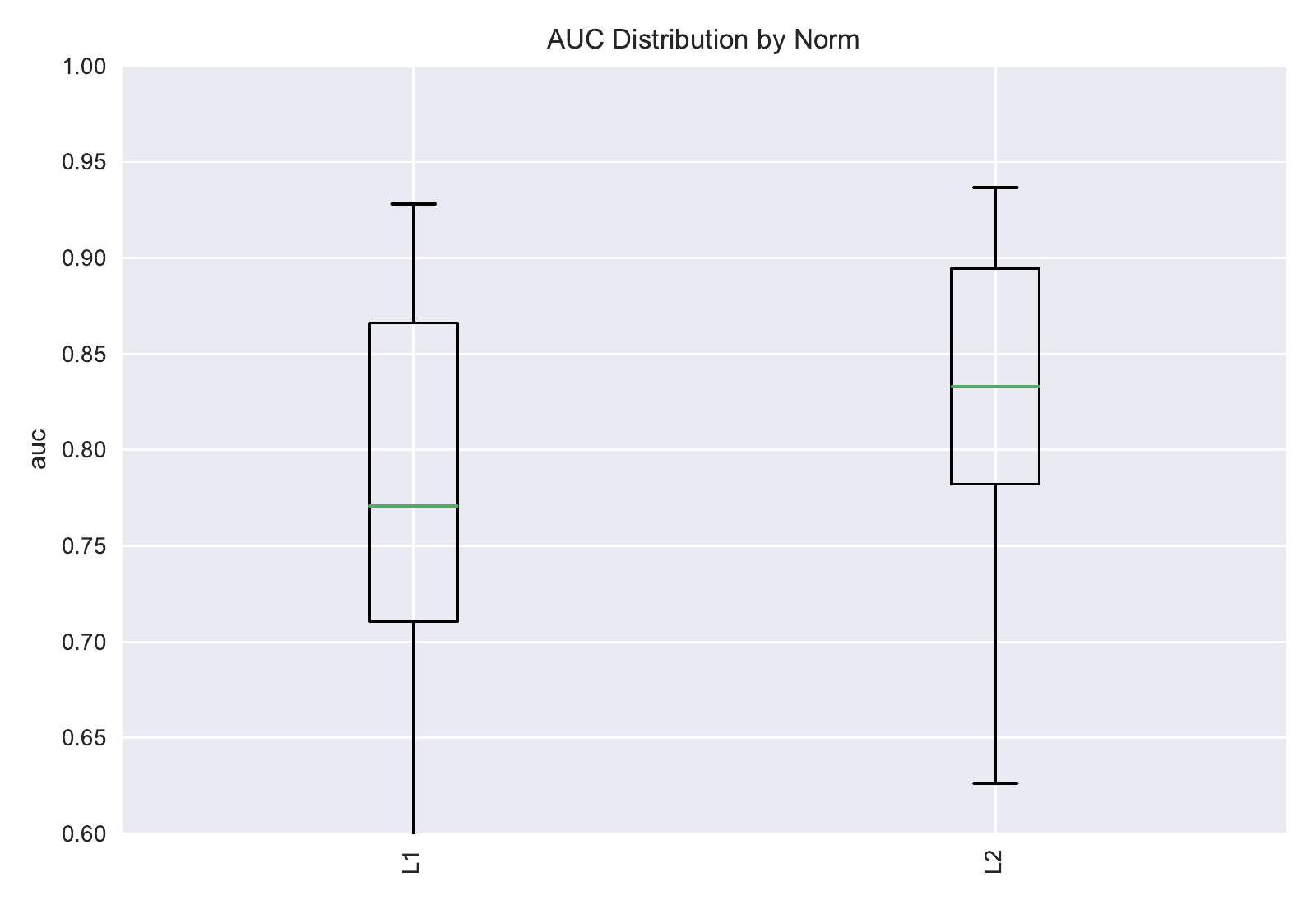}
\includegraphics[width=0.48\columnwidth]{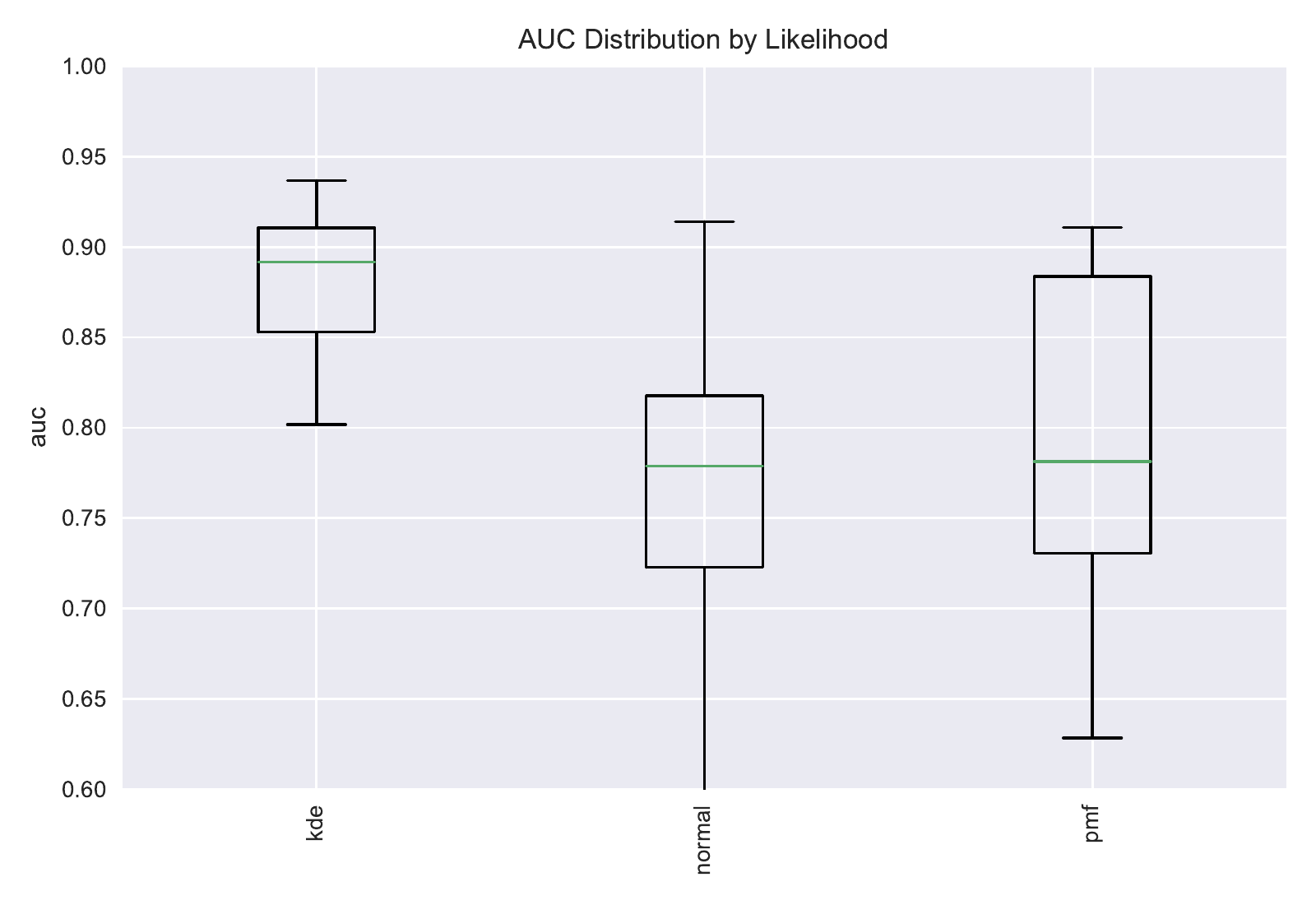}
\includegraphics[width=0.48\columnwidth]{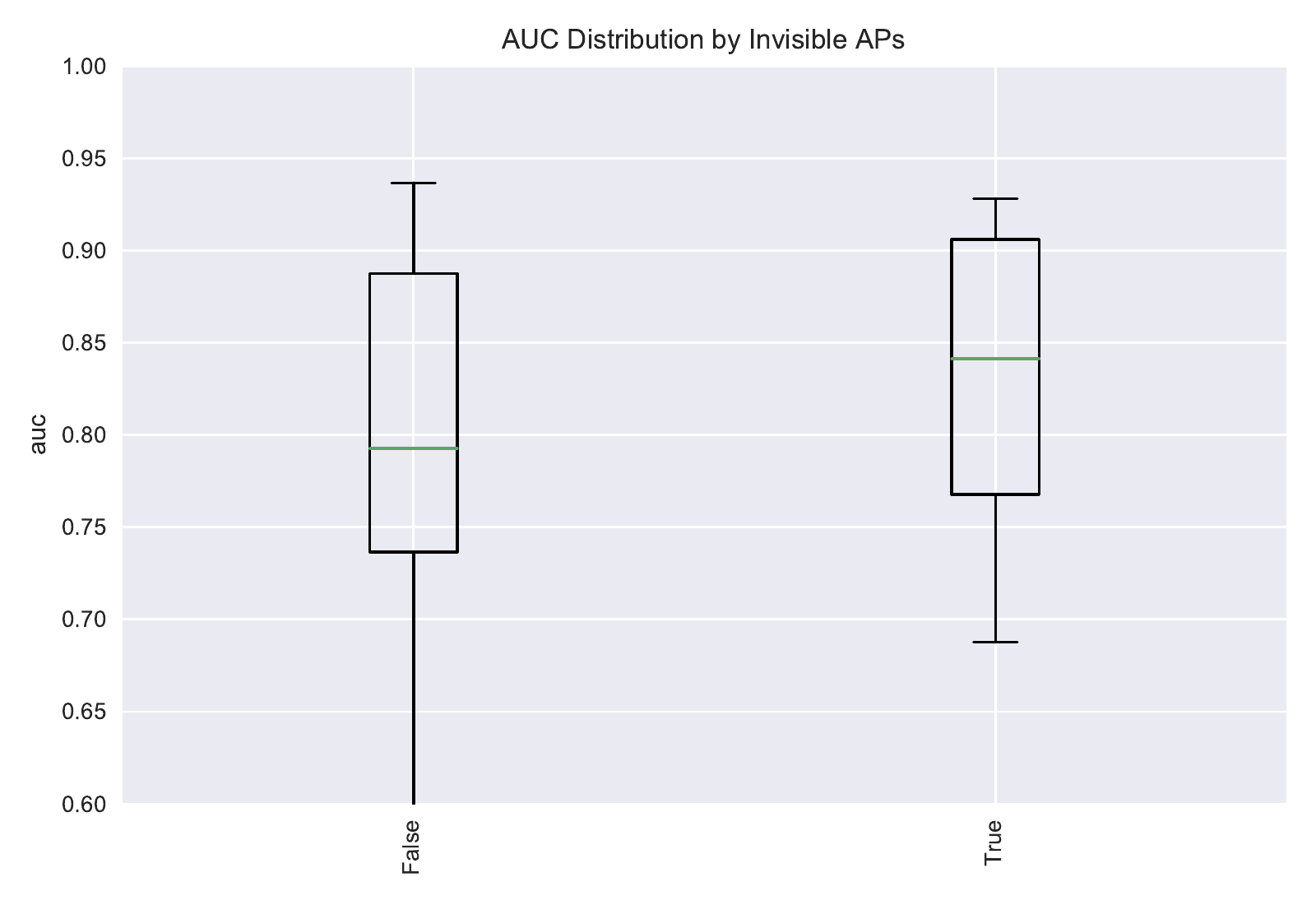}



%
\caption{Distribution of calculated AUC values for various groups.}\label{fig:comp_dist_measure}
\end{centering}
\end{figure}
Figure \ref{fig:comp_dist_measure} shows summary statistics of the calculated AUC values for various distance measures. A surprising result to us is that simple absolute distances between expected values of the distributions often gave the best performance. Our actual intuition was that a measure, which does not consider the subtle differences between distributions, could not capture these distances well. Therefore, our actual intention was to include this measure as a baseline.

Another surprising result was that EMD often gave almost the same
results as the absolute distance between the means and therefore
performed about equally well. This seems obvious to us
and can be explained as follows: Consider two univariate CDFs. For
EMD, we calculate the area between both curves. Now consider the case
where each CDF jumps from $0$ to $1$ in a small interval where the
measured RSSI-values are located. Let the positions of these intervals
be located far away from each other. Then the area between both curves
is approximately proportionate to the distance between those small
intervals. Therefore, EMD is often dominated by the absolute
difference of the mean RSSI
values. 
We expected EMD to perform well because of its properties, e.g., it
fulfills the triangle inequality. Therefore, it coincides with our
intuition that there cannot be any shortcut to the direct path between
two locations over a third location. It also incorporates the distance
between non-overlapping distributions and provides smoothing through
the CDF. Therefore it was no surprise to us, that the calculated
distances are  stable across different likelihood
estimations, which however can also be attributed to the
mentioned relation to the absolute difference of means. An
advantage of this measure is that it already gives good
results with a likelihood representation obtained through a simple
counting of RSSI values. We think an advantage of EMD over the
absolute differences between means is that it can also capture subtle
differences between close distributions, e.g., when their means are
similar but their variances or kurtoses differ. For more distant
distributions it gives very similar results to the distances between
means, which seems a positive property.

Concerning the other distance measures, we sometimes observed specific
settings, where similar or even slightly better results could be
achieved for some smartphones. However, these measures are a lot more
sensitive to the likelihood estimation and the used norm function. We
think that stability, and therefore reliability of the used measure,
should be preferred.

\subsubsection{Euclidean Norm vs. Manhattan Norm}
Almost all distance measures profited from calculating the $L_2$ norm
instead of the $L_{1}$ norm of the distance vector. In particular, ROC
curves of worse performing distance measures improved. Although, well
performing measures improved only slightly, or did perform weaker. We
believe that through the Euclidean norm our distance measure becomes
more sensitive to individual greater distances between the univariate
distributions. Those distance measures that are bounded by a maximum
value, i.e., the distributions do not overlap, and do not
differentiate between larger or smaller gaps between the
distributions, therefore profit from the $L_2$ norm. Nonetheless, we found evidence in
literature that for clustering high-dimensional data the $L_1$-norm
has advantages \cite{keimhighdim}. However, this does not
seem to hold for our data. The reason for this may be found in the
fact that we basically compute distances in lower-dimensional sub spaces.

\subsubsection{Likelihood Estimation}
From the estimation techniques, KDE consistently worked better than
using a PMF. We attribute this to the smoother distributions obtained.
Therefore they are more robust to the effects of random sampling. Normal distributions gave the lowest performance in our experiments. From a theoretical point of view, it makes little sense to estimate a normal distribution, when AP invisibility should be modeled. This is because such distributions would need to be multimodal. 

\subsubsection{Modelling AP Invisibility}
Representing AP invisibility in the distributions gave strong
improvements in most cases, especially in combination with KDE.  In
\cite{ref_kl}, localization is achieved only through modelling the
multinomials of AP visibilities. We think that representing both, AP
invisibilities and RSSI values, is a strong method for location
discrimination. 


\section{Summary and Conclusions}
In this work we studied various distance measures and representations
of WiFi observations. After our investigation we are able to provide
the following \textsl{rules of thumb} as recommendations for measuring
dissimilarities between RSSI likelihoods. As it turned, applying
kernel density estimation for the likelihoods is the method of
choice. Also, one should include AP invisibility into the modeling.
Finally, the Earth Mover's Distance can deliver subtle differences for
close distributions as well as stable differences for far
distributions.


We also unraveled several limitation of our approach.  Since all of
our observations were measured on the same floor of the building, the
results may not be transferable to three dimensional
scenarios. However, we believe that the effect of floors and ceilings
between vertically stacked rooms is similar as the effect of walls
between neighbored rooms. This should be verified in future work.  The
employed approach for motion mode segmentation may fail whenever
artificial acceleration patterns appear, e.g., using an
elevator. Nonetheless, utilizing a more elaborate activity recognition
system (e.g., \cite{ref_wardrive}) would be able to improve this
situation.

Concluding we would like to point out, that our ultimate goal is
beyond a localization through distances alone. We rather consider our
investigation of distance measures as a building block for more
complex localization techniques. Hence, we investigated in this work
the properties of this building block. These techniques may not
necessarily depend on WiFi. While we were aiming in this work at
clustering WiFi distributions, we are convinced our results can be
transferred to other scenarios. For example, they may be also
applicable in localization scenarios employing Bluetooth low energy
(BLE) beacons.

\bibliography{mybib}{}
\bibliographystyle{plain}


\end{document}